\documentclass{article} 
\usepackage{iclr2027_conference,times}

\usepackage{amsmath,amsfonts,bm}

\def\eqref#1{equation~\ref{#1}}

\def\1{\bm{1}}

\DeclareMathAlphabet{\mathsfit}{\encodingdefault}{\sfdefault}{m}{sl}
\SetMathAlphabet{\mathsfit}{bold}{\encodingdefault}{\sfdefault}{bx}{n}

\usepackage{graphicx}
\usepackage{booktabs}
\usepackage{multirow}    
\usepackage{adjustbox}   

\usepackage{hyperref}

\newenvironment{fittable}
  {\begin{adjustbox}{max width=\linewidth,center}}
  {\end{adjustbox}}

\newcommand{\side}[2]{\multirow{#1}{*}{\rotatebox[origin=c]{90}{#2}}}
\newcommand{\sidecite}[2]{\multirow{#1}{*}{\rotatebox[origin=c]{90}{\scriptsize #2}}}

\newcommand{\method}{TEX-MEX}

\newcommand{\dec}[1]{\textcolor{red!75!black}{$-$#1}}
\newcommand{\inc}[1]{\textcolor{green!50!black}{$+$#1}}

\title{\method{}: Similarity-Guided Temporal\\ Contrastive Learning for Time Series}

\author{Aditya Radhakrishnan$^{1}$, Maxwell A. Xu$^{1}$\thanks{Also
affiliated with Google.}\,, Wanting Mao$^{1}$, Santosh Kumar$^{2}$, \\
\textbf{Harish Haresamudram$^{1}$, James M. Rehg$^{1}$} \\
$^{1}$Health Care Engineering Systems Center, University of Illinois Urbana-Champaign \\
$^{2}$Department of Computer Science, University of Memphis \\
\texttt{\{aditya63, jrehg\}@illinois.edu}
}

\iclrfinalcopy

\begin{document}

\maketitle

\begin{abstract}
Contrastive learning for time series depends on deciding which pairs of inputs should be treated as similar and dissimilar. 
Because few augmentations reliably preserve meaning in time series, prior methods have relied on two disjoint notions of similarity between windows: structural similarity, which relates windows with similar morphology, and temporal similarity, which relates windows that occur close together in time. 
We introduce \method{}, to our knowledge the first approach to unify the two into a single relationship. 
Given a random batch of windows drawn across the dataset, \method{} uses the encoder to identify which windows are structurally similar or dissimilar to each other. 
It then pulls each window toward the temporally shifted versions of similar windows and pushes it away from those of dissimilar ones, learning what typically accompanies similar windows in time. 
Across PPG, EEG, ECG, accelerometry, and continuous glucose monitoring, \method{} broadly achieves state-of-the-art performance against contrastive, reconstructive, JEPA-based, and modality-specific approaches, establishing it as a strong general SSL objective. 
We further train a 560K-parameter PPG foundation model that achieves state-of-the-art performance across a broad range of downstream benchmarks while being roughly 50$\times$ smaller than the next-best unimodally pretrained PPG foundation model in our evaluation. We will release this foundation model open-source and open-weight to the community to support future work and practical applications.
\end{abstract}

\section{Introduction}
\label{sec:intro}

Self-supervised learning (SSL) has emerged as the dominant paradigm for learning generalizable representations from large-scale unlabeled data \citep{chen2020simple, he2022masked}. This approach is particularly attractive for modeling physiological signals, such as photoplethysmography (PPG) or electrocardiography (ECG), which are routinely captured as time series data by wearables (e.g. smart watches) or medical devices. Labeling clinically-important events in such data (such as atrial fibrillation in an ECG recording) typically requires expert interpretation, invasive assays, or controlled clinical settings \citep{xu2024rebar}, and is very difficult to perform at scale. At the same time, the increasing adoption of smart watches by the general public and the large-scale aggregation of  waveform data within electronic health records has created an opportunity to develop foundation models for such physiological signals \citep{xu2025relcon, saha2025pulse, ansari2025chronos}. Such models could enable new data-driven approaches to modeling states of health and disease.

Time series representation learning is challenging due to the need to capture statistical dependencies across widely-varying temporal scales \citep{goldberger2002fractal}. For example, heart rate, which is measurable via PPG, is correlated at the level of minutes in responding to physiological demands such as exercise, but is also correlated at the level of hours due to the diurnal cycle. Weekly patterns of stress or physical activity, such as regular commute times or exercise routines, can induce even longer-range correlational structure. A key step in developing foundation models for such data is the identification of effective pretext tasks. Two common approaches are contrastive learning, which extracts windows of samples from a time series recording to construct positive and negative pairs, and masked autoencoding, which uses temporal context to reconstruct the samples within a masked-out window. Contrastive learning is a particularly useful lens for examining time series representation learning as it provides a way to explicitly reason about similarity and dissimilarity. 

\begin{figure}
    \centering
    \includegraphics[width=0.85\linewidth]{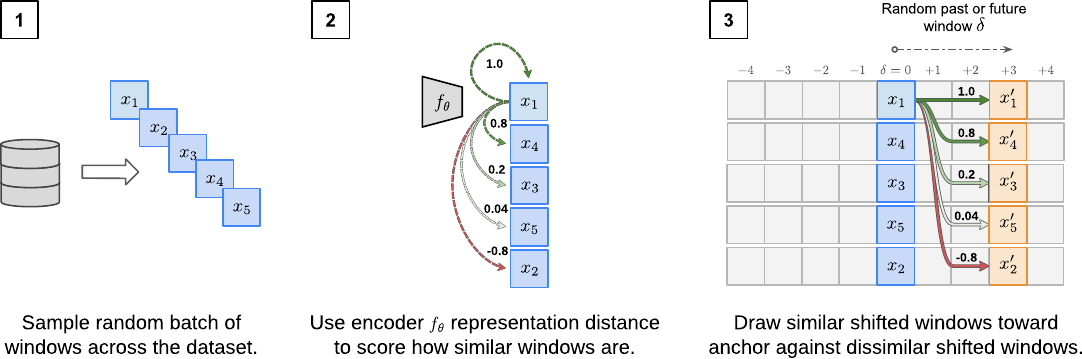}
    \caption{\textbf{Overview of TEX-MEX.} (1) Windows of sensor data are randomly sampled from the dataset;
    (2) The encoder scores each window's similarity to an anchor (here, $x_1$) using cosine similarity; 
    (3) Each window is offset in time (forward or backward) by a random  $\delta$ (here, $+3$ windows). 
    The anchor is pulled toward the temporal shifts of similar windows and away from those of dissimilar ones, in proportion to the similarity, composing structural and temporal notions.
    }
    \label{fig:teaser}
\end{figure}

Unfortunately, time series recordings typically lack a rich set of well-defined invariances from which positive pairs can be constructed~\citep{xu2024rebar}. 
As a consequence, prior contrastive learning works have adopted one of two disjoint approaches to defining similarity between windows: \emph{structural similarity} approaches that index on signal morphology or frequency content (i.e., windows are similar if they contain similar waveform shapes, like two heart beats in an ECG waveform) \citep{xu2025relcon, xu2024rebar}, and \emph{temporal similarity} measures that index on temporal proximity (i.e., two windows are similar if they occur close together in time, such as when a cool-down period follows a period of vigorous exercise) \citep{tonekaboni2021unsupervised, oord2018representation}. 
In contrast, this paper proposes, to our knowledge, the first contrastive learning approach which unifies structural and temporal similarity measures into a single framework. 
We demonstrate that the standard approach of generating positive pairs by mining morphologically-similar patterns from time series recordings \citep{xu2024rebar} can be extended by composing it with temporal similarity, drawing positive and negative pairs from the temporal context of similar and dissimilar windows. 
This unifies the two primary sources of similarity in time series modeling: windows that correspond to similar physiological processes (e.g., two heartbeats) will have similar morphology, while windows which map onto stable patterns of human behavior (e.g., cool-down following exercise) will be temporally co-occurring. 
An exercise window, for instance, can learn from the cool-downs that follow exercise in other recordings, capturing what typically follows exercise rather than what followed it once.

We introduce \method{}: for each anchor window, it first identifies similar and dissimilar windows within a random batch drawn across the dataset. 
Every window is then shifted by the same random offset, forward or backward in time, redrawn in each batch to span many timescales. 
\method{} pulls the anchor toward the shifted versions of similar windows and pushes it away from those of dissimilar windows. 
We find that the encoder's own representations suffice for measuring this similarity, eliminating the costly signal distance measures of prior methods \citep{xu2024rebar}. 
Unlike those fixed measures, this similarity improves as the encoder trains, ultimately surpassing them. 
Because it is graded rather than binary, \method{} is naturally a soft contrastive method \citep{lee2024soft, xu2025relcon}, weighting each shifted window by its source window's similarity to the anchor, which captures degrees of relatedness and adds robustness to false negatives. 
The main contributions of our work are as follows:

\begin{enumerate}
    \item We introduce \method{}, a contrastive objective that unifies structural and temporal similarity by contrasting the shifted versions of similar windows against those of dissimilar windows, with similarity derived from the encoder's own representations.
    \item We show that this representation-space similarity improves throughout training, surpassing existing signal distance approaches as a predictor of mutual class membership.
    \item We demonstrate the generality of \method{} across PPG, ECG, accelerometry, EEG, and CGM, where it broadly achieves state-of-the-art performance against leading contrastive, reconstructive, JEPA-based, and modality-specific methods.
    \item Leveraging its efficiency, we train and will release an open-source, open-weight PPG foundation model that achieves state-of-the-art performance across a range of downstream tasks with just 560K parameters, roughly 50$\times$ smaller than the next-best unimodally pretrained PPG foundation model in our evaluation.
\end{enumerate}

\section{Related Work}
\label{sec:related}

\textbf{Time-Series Self-Supervised Learning.} Temporal contrastive methods use proximity in time as supervision: CPC \citep{oord2018representation} contrasts a window's future against other samples, and TNC \citep{tonekaboni2021unsupervised} treats windows within a locally stationary neighborhood as positives, reweighting others as unlabeled. A single neighborhood, however, ties relatedness to one timescale and one recording. Structural methods instead relate windows by learned distances, such as REBAR's reconstruction-based motif similarity \citep{xu2024rebar} and RelCon's learnable distance \citep{xu2025relcon}, which RelCon also uses to grade relationships for soft contrastive learning, as in our work. These, however, miss temporal relationships, relating only windows that look alike. TS2Vec \citep{yue2022ts2vec} and SoftCLT \citep{lee2024soft} apply instance-wise and temporal contrast as losses independent of one another, and their temporal contrast stays within a single input. SoftCLT softens both, using DTW distance and a monotonically decaying timestamp distance, and reports that embedding-space similarity, as in NNCLR \citep{dwibedi2021little}, degrades performance; in our formulation, where it determines how temporal context is contrasted rather than serving directly as positives, it instead surpasses data-space distances. Masked modeling approaches such as TimeMAE \citep{cheng2026timemae} and PatchTST \citep{nie2022time} instead reconstruct masked portions of an input.

\textbf{PPG Foundation Models.} Recent PPG foundation models span contrastive, reconstructive, and autoregressive pretraining. Contrastive approaches differ mainly in how they relate windows: through shared subject identity~\citep{abbaspourazad2023large, pillai2025papagei}, morphology-based similarity~\citep{pillai2025papagei, saha2025pulse}, quality-aware temporal proximity~\citep{Ding2024SiamQuality}, or synchronization with paired ECG~\citep{nie2026anyppg}. xMAE~\citep{zhou2026xMAE} instead learns through cross-modal reconstruction of masked ECG, and GPT-PPG~\citep{chen2025gpt} models PPG autoregressively through next-patch prediction. We build on Pulse-PPG~\citep{saha2025pulse}, adopting its pretraining corpus and downstream evaluation.

\section{Methodology}
\label{sec:method}

\paragraph{Batch Construction}

At each training step, we randomly sample a batch of $B$ windows, $\mathcal{X}=\{x_1,\ldots,x_B\}$, across the pretraining dataset. 
Then, for each batch, we sample a single temporal offset 
$\delta \sim \mathcal{N}(0,\sigma^2)$, measured in window lengths. 
Unlike in TNC, $\delta$ does not determine positives: it is shared 
across the batch, and each shifted window's role as positive or 
negative is set by encoder-estimated similarity, as described next. 
The offset thus only selects the timescale each batch probes, and the 
Gaussian acts as a simple, symmetric prior over where useful relationships are likely 
to be exposed, with $\sigma$ setting their scale.
We reject and resample draws outside 
$\delta_{\min} \le |\delta| \le \delta_{\max}$. The lower bound prevents shifted windows from overlapping their originals, which would permit trivial identity shortcuts; the upper bound limits 
shifts that cross gaps or recording boundaries. We set $\delta_{\min}=1$ and $\delta_{\max}=2.5\sigma$ by default, the latter truncating only ${\approx}1.2\%$ of the Gaussian's mass.
Each window $x_i$ is then paired with a shifted offset (a past or future window) $x_i'=T_\delta(x_i)$ from the same recording, where $T_\delta$ denotes displacement by $\delta$ along the time series. 
This produces paired unshifted and shifted sets, $\mathcal{X}=\{x_1,\ldots,x_B\}$ and $\mathcal{X}'=\{x'_1,\ldots,x'_B\}$, respectively.

\begin{figure}
    \centering
    \includegraphics[width=\linewidth]{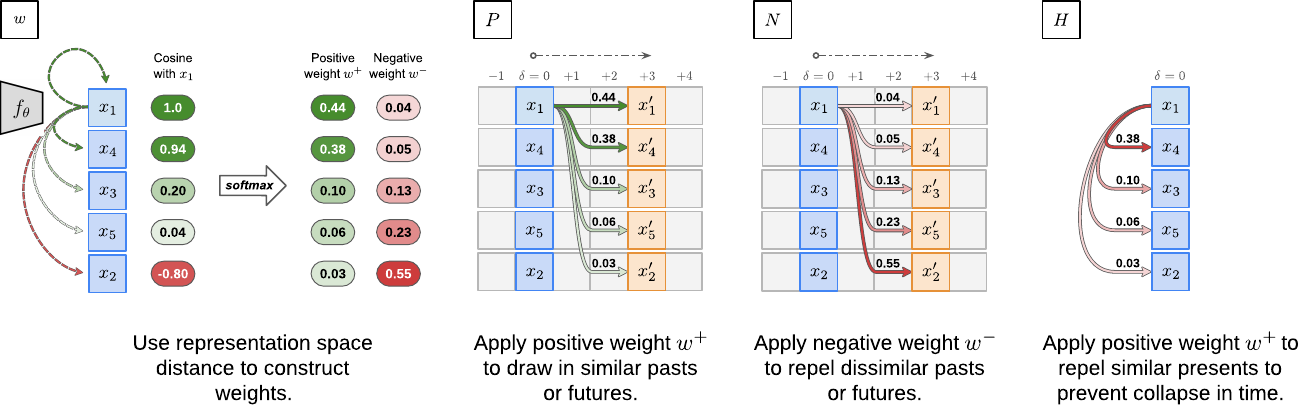}
    \caption{\textbf{Components of the TEX-MEX loss}, shown for anchor $x_1$ with a batch of five windows and $\delta = +3$. ($w$) The anchor's cosine similarity to each window is turned by two softmaxes into positive weights $w^+$, which favor similar windows, and negative weights $w^-$, which favor dissimilar ones. ($P$) The shifted windows, weighted by $w^+$, are attracted to the anchor. ($N$) The same shifted windows, weighted by $w^-$, are repelled. Every shifted window therefore counts as partly positive and partly negative, according to how related it is. ($H$) The unshifted versions of the other windows, weighted by $w^+$, are also repelled. This stops the encoder from ignoring time by mapping a window and its shift to the same point. The loss is $-\log \bigl( P / (P + N + H) \bigr)$.}
    \label{fig:method}
\end{figure}

\paragraph{Finding Similar and Dissimilar Windows: Encoder-Based Batch Grading}
A useful property of \method{} is the availability of continually improving estimates of the degree of similarity between windows, derived directly from the encoder's representation space (as examined in Section~\ref{sec:pulse}). 
Let $z_i=f_\theta(x_i)$ denote the $\ell_2$-normalized representation of window $x_i$.
For each anchor $x_i$, we compute its cosine similarity with every member $x_k$ of the unshifted batch as $c_{ik}=z_i^\top z_k$. 
We then transform these similarities into positive and negative weights,
\begin{equation}
w_{ik}^{+}=\frac{\exp(c_{ik}/\tau_w)}{\sum_{j=1}^{B}\exp(c_{ij}/\tau_w)} \qquad w_{ik}^{-}=\frac{\exp(-c_{ik}/\tau_w)}{\sum_{j=1}^{B}\exp(-c_{ij}/\tau_w)}
\end{equation}
where $\tau_w$ controls the sharpness of the weighting distributions. 
Thus, $w_{ik}^{+}$ places greater weight on batch members most similar to the anchor, while $w_{ik}^{-}$ places greater weight on those least similar. Together, these provide complementary estimates of the degree of similarity between each anchor and all $B$ members of the batch. When producing these weights, we apply a stop-gradient to $c_{ik}$, preventing the contrastive objective described below from directly optimizing the similarity estimates used to construct its own weights. However, because explicit negatives remain, removing the stop-gradient does not strictly result in collapse.

\paragraph{Soft Contrastive Terms on a Graded, Temporally-Shifted Batch}

Having estimated similarity among the unshifted windows, we apply these relationships to attract and repel shifted windows in time. Let $z'_k=f_\theta(x'_k)$ denote the normalized representation of shifted window $x'_k$. For each anchor $x_i$, we compute its similarity to every shifted batch member as $s'_{ik}=z_i^\top z'_k/\tau$, where $\tau$ is the contrastive temperature. We then use the weights derived above to construct positive and negative terms,
\begin{equation}
P_i=\sum_{k=1}^{B}w^+_{ik}\exp(s'_{ik}) \qquad N_i=\sum_{k=1}^{B}w^-_{ik}\exp(s'_{ik})
\end{equation}
Thus, the same set of shifted candidates participates in both terms, but their contributions differ according to their estimated relationship with the anchor: shifted windows of more similar members contribute more strongly to $P_i$, while those of more dissimilar members contribute more strongly to $N_i$. In this way, \method{} contrasts the temporal shifts of batch members according to their graded cross-member relationships rather than temporal distance.

\paragraph{Encouraging Temporal Frame Separation}

The positive and negative terms above contrast relationships across batch members in time, but alone do not require representations to preserve features that distinguish a window from its temporal shift: an encoder could merely learn a collapsed invariance to time. For example, user identity shortcuts may be learned. To prevent this, we additionally contrast the temporal shifts of similar batch members against their corresponding unshifted windows. We define $s_{ik}=z_i^\top z_k/\tau$ and introduce the temporal frame-separation term \begin{equation} H_i=\sum_{k\neq i} w^+_{ik}\exp(s_{ik}) \end{equation} Here, unshifted windows of more similar batch members receive greater weight as negatives. We exclude $k=i$, as a sample cannot serve as a negative against itself. Consequently, for a similar member $k$, the shifted representation $z'_k$ contributes to the positive term $P_i$, while its unshifted representation $z_k$ contributes with the same  weight $w^+$ to the negative term $H_i$, encouraging separation between temporal frames.

Combining the three terms, the \method{} objective is \begin{equation} \mathcal{L}_{\text{\method{}}}=-\frac{1}{B}\sum_{i=1}^{B}\log\frac{P_i}{P_i+N_i+H_i} \end{equation} The resulting objective attracts an anchor toward the temporal shifts of similar batch members, repels it from the temporal shifts of dissimilar members, and distinguishes those shifted representations from the original unshifted frame: \underline{te}mporal \underline{cross}-\underline{m}ember \underline{ex}change.

\section{\method{} as a General Time-Series SSL Objective}
\label{sec:general}

To evaluate \method{} as a general time-series SSL objective, we adopt two recently proposed evaluation settings spanning five signal domains: first, following PULSE \citep{chen2025self} (PPG, EEG, ECG, and HAR) and second, following CGM-JEPA \citep{muhammad2026cgm} (for CGM).

\subsection{Multi-Domain Self-Supervised Learning: PPG, EEG, ECG, and HAR}
\label{sec:pulse}

\paragraph{Setup}

The core evaluation setting of PULSE \citep{chen2025self} utilizes four datasets spanning four time-series domains. For PPG, WESAD \citep{schmidt2018introducing} is used for four-class affective-state classification across baseline, stress, amusement, and meditation states. For EEG, Sleep-EDF \citep{kemp2000analysis} is used for five-class sleep-stage classification. For ECG, MIT-BIH Atrial Fibrillation \citep{moody1983new} is used for binary classification of atrial fibrillation versus normal rhythm. Finally, the accelerometry dataset from \cite{reyes2015smartphone} is used for six-class human activity recognition. We retain preprocessing, windowing, and predefined 70/15/15 inter-subject splits.

The PULSE evaluation protocol provides a standardized setup for introducing new SSL objectives while preserving the encoder architecture and downstream evaluation. We use the same 10-layer dilated-convolution encoder with 320-dimensional representations. Each method is tuned through 30 random-search trials over shared training hyperparameters, while allowing method-specific hyperparameters to vary independently. The best hyperparameter configuration is selected by validation performance, and the lowest-validation-loss checkpoint within each training run is retained. Frozen representations are evaluated using a fixed logistic-regression linear probe, with results averaged across five model initialization seeds (0--4) on the predefined split. We reuse results for existing baselines from \citet{chen2025self}, which cover representative modern approaches spanning contrastive learning, masked autoencoding, and sequential variational autoencoding.

Alongside \method{}, we implement TNC \citep{tonekaboni2021unsupervised} and SoftCLT \citep{lee2024soft} as additional baselines representing prior temporal contrastive approaches. For each, we sweep method-specific hyperparameters following the search spaces of their original implementations. Further details on hyperparameter sweeps and final choices can be found in Appendix~\ref{app:pulse_hparams}.

\paragraph{\method{} as a Distance Function is a Predictor of Mutual Class
Membership}

\begin{figure}[t]
\centering
\includegraphics[width=\linewidth]{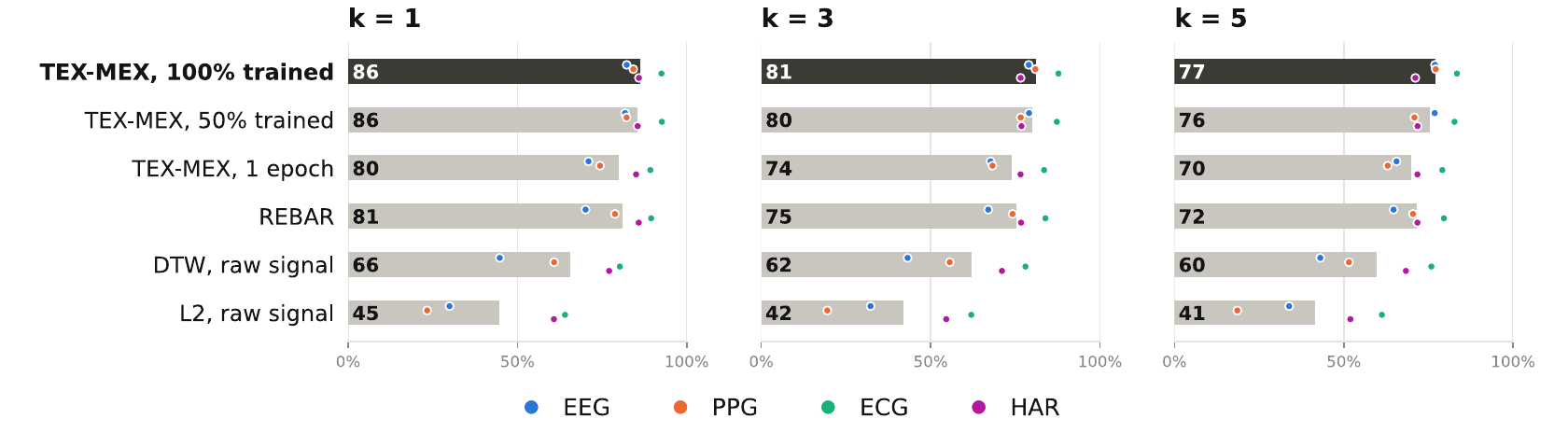}
\caption{\textbf{Comparison of signal distance measures by mutual class membership.} On the held-out set, we sample 100 batches of 64 windows per domain and, for each anchor, retrieve its $k\in\{1,3,5\}$ nearest neighbors according to each distance measure, reporting the proportion sharing the anchor's ground-truth class. Bars show the unweighted mean across PPG, EEG, ECG, and HAR, with colored points indicating individual domains. \method{} is evaluated at 1 epoch, 50\% of training, and after training, and compared with REBAR and raw-signal DTW and $L_2$ distances. Higher values indicate that a distance measure more reliably identifies semantically similar signals.}
\label{fig:mutualmem}
\end{figure}

The use of signal-distance functions to estimate similarity has been applied in several prior time-series SSL works. REBAR \citep{xu2024rebar} learns a reconstruction-based distance that captures shared motifs between windows, while SoftCLT \citep{lee2024soft} uses DTW in the raw signal space to estimate graded relationships between instances. We derive the former from \cite{chen2025self}. Figure~\ref{fig:mutualmem} compares these approaches with the representation-space distance learned directly by \method{}, as well as raw-signal $L_2$ distance. For each domain, we sample 100 batches of 64 windows from the held-out set and measure the proportion of each anchor's $k\in\{1,3,5\}$ nearest neighbors that share its ground-truth class. All distance measures are evaluated on identical batches, and we evaluate \method{} at one epoch, halfway through training, and after training.

A few observations emerge from this analysis. First, by 50\% of training, the distance learned by \method{} is the strongest measure across all three values of $k$, exceeding REBAR by 4--5 percentage points on average. Second, its continuous improvement suggests that the representation space becomes an increasingly informative source of cross-member relationships as optimization proceeds. Unlike REBAR or DTW, for which the distance measure is frozen for contrastive learning, there is no fixed ceiling in supervision quality.

\paragraph{Results} 

\begin{table}[t]
\setlength{\tabcolsep}{4pt}
\renewcommand{\arraystretch}{0.9}
\caption{Linear probe performance across four sensor domains. \textbf{Bold} is
best, \underline{underline} second best. The bracketed block is reproduced from
\citet{chen2025self}; TNC and SoftCLT are run by us under the same protocol.}
\label{tab:pulse}
\begin{fittable}
\footnotesize
\begin{tabular}{clcccccccc}
\toprule
& & \multicolumn{4}{c}{AUROC $\uparrow$} & \multicolumn{4}{c}{PRAUC $\uparrow$} \\
\cmidrule(lr){3-6}\cmidrule(lr){7-10}
& Method & PPG & ECG & HAR & EEG & PPG & ECG & HAR & EEG \\
\midrule
\side{8}{\citet{chen2025self}}
  & SimCLR \citep{chen2020simple}   & .6119 & .8254 & .9938 & .8545 & .3608 & .8063 & .9763 & .5095 \\
  & TS2Vec \citep{yue2022ts2vec}  & .6428 & .8656 & .9931 & .9499 & .3959 & .8516 & .9766 & .7022 \\
  & REBAR \citep{xu2024rebar}    & .6977 & .9146 & \underline{.9965} & .9508 & .4457 & \underline{.8985} & \underline{.9891} & .7077 \\
  & PatchTST \citep{nie2022time} & .7108 & .7096 & .9744 & .9355 & .5291 & .6816 & .8990 & .6737 \\
  & TimeMAE \citep{cheng2026timemae}  & .7808 & .7661 & .9914 & .9509 & .5674 & .7662 & .9705 & .7322 \\
  & LFADS \citep{sussillo2016lfads}    & .7110 & .7169 & .9949 & .9449 & .4959 & .6921 & .9829 & .6855 \\
  & DSVAE \citep{yingzhen2018disentangled}    & .7678 & .8288 & .9936 & .9542 & .5538 & .8131 & .9769 & .7225 \\
  & PULSE \citep{chen2025self}    & \underline{.8029} & \underline{.9493} & .9942 & .9617 & \underline{.5989} & \textbf{.9475} & .9810 & .7382 \\
\midrule
& TNC \citep{tonekaboni2021unsupervised}        & .7213 & .8723 & .9923 & \underline{.9628} & .5016 & .8421 & .9734 & \underline{.7459} \\
& SoftCLT \citep{lee2024soft}    & .7089 & .8598 & \textbf{.9969} & .9527 & .4550 & .8118 & \textbf{.9893} & .7170 \\
\midrule
& \method{}   & \textbf{.8162} & \textbf{.9539} & .9944 & \textbf{.9684} & \textbf{.6094} & \textbf{.9475} & .9811 & \textbf{.7580} \\
\bottomrule
\end{tabular}
\end{fittable}
\end{table}

Table~\ref{tab:pulse} summarizes linear-probe performance across the four domains. For PPG, EEG, and ECG, \method{} achieves the highest (or, in one case, tied-highest) performance on both AUROC and PRAUC, with PULSE generally providing the closest baseline performance. Further, \method{} retains its computational efficiency: on a Quadro RTX 6000, a controlled runtime benchmark found that \method{} reduced peak GPU memory utilization by 65.4\% and pretraining wall-clock time by 49.6\% on average relative to PULSE. On HAR, however, \method{} is outperformed by SoftCLT, LFADS, and REBAR, ranking 4/11. Nevertheless, performance on this task is highly saturated: all but one baseline exceeds 99\% AUROC, with \method{} reaching 99.44\% compared with 99.69\% for SoftCLT, the best-performing method. A similar, though somewhat larger, gap is observed for PRAUC, where \method{} reaches 98.11\% compared with 98.93\%.

The advantage of \method{} is especially clear in comparison with contrastive methods that similarly exploit temporal structure. Across the benchmark, \method{} is the only such approach that remains consistently competitive with the strongest SSL methods, with particularly large gains on PPG and ECG over TNC, SoftCLT, and the SimCLR implementation of \citet{chen2025self}, which uses temporal shifting as an augmentation. Each of these methods prescribes similarity in advance, whether by temporal proximity (TNC, SimCLR) or by DTW and timestamp distance (SoftCLT). \method{} instead learns similarity through the encoder itself, contrasting the temporal shifts of similar windows against those of dissimilar ones. These learned relationships carry most of the positive weight, with only 30.2\% of $w^+$ (averaged across domains and epochs) falling on the anchor's own shift. Because similarity comes from representations already computed in the forward pass, \method{} also avoids the highly costly dataset-wide preprocessing required by TNC (ADF stationarity testing) and SoftCLT (DTW distances), making it substantially more amenable to large-scale pretraining. We exploit this advantage of \method{} in Section~\ref{sec:fm}.

\subsection{\method{} Applied to Continuous Glucose Monitoring}
\label{sec:cgm}

\paragraph{Setup} We follow the experimental setting of CGM-JEPA \citep{muhammad2026cgm}, preserving its transformer-based encoder, pretraining data, downstream tasks, seeds, and evaluation protocol. Pretraining uses unlabeled continuous glucose monitoring (CGM) recordings from a Stanford study of metabolic health \citep{metwally2025prediction} together with a public CGM dataset \citep{colas2019detrended}. Downstream evaluation uses two labeled groups from the Stanford study. One group contains glucose responses measured from venous blood draws during an in-clinic oral glucose tolerance test (OGTT), while a separate held-out group contains both venous OGTT measurements and wearable CGM collected at home. Each participant is clinically labeled for insulin resistance and $\beta$-cell dysfunction, yielding two binary prediction tasks. We evaluate frozen representations in three settings: training on venous measurements from the first group and testing on venous measurements from the held-out group, training and testing on home CGM within the held-out group, and training on venous measurements and testing on home CGM within the held-out group. The logistic-regression probe and average of 20 seeds of subject-level 2-fold cross validation are unchanged, and all baseline results (other than TNC and SoftCLT) are derived from \citet{muhammad2026cgm}.

Incorporating \method{} requires  a few tweaks in this setup. First, the corpus does not support temporal shifts at CGM-JEPA's 24-hour (288-reading) window: 391 of its 413 recordings are exactly one day long, leaving only 22 long enough to hold both an anchor and a displaced partner. We therefore rewindow and pretrain on 39-reading (3.25-hour) windows, matching the 39-point curves that form every downstream input in their setup. Second, we z-score inputs using corpus statistics; this is applied to TS2Vec in their setup. We derive all \method{}-specific hyperparameters without modification from defaults derived in Section~\ref{sec:fm}, with an exception for $\sigma=1$ chosen beforehand as a more plausible timescale for the domain. We also expand the baselines list by including TNC and SoftCLT. For the former, we use the \method{} window size of 39, while for the latter, we preserve the window size of 288 as SoftCLT only performs within-window temporal contrast. For both, we use defaults for method-specific hyperparameters.

\paragraph{Results} Table~\ref{tab:cgm} summarizes linear-probe performance for metabolic subphenotyping. \method{} achieves the best or tied-best performance on 11 of 12 metric-task combinations, spanning both $\beta$-cell dysfunction and insulin resistance across the three evaluation settings. The sole exception is venous insulin resistance AUROC, where TNC achieves the highest performance.

The largest gains emerge on at-home CGM, particularly for insulin resistance. \method{} exceeds the strongest competing baseline by 6.3 AUROC points for in-domain home evaluation and by 8.6 points under venous$\rightarrow$home transfer. The latter is the largest improvement across the evaluated settings and is particularly relevant for practical deployment, as it tests whether a probe learned from clinically collected venous OGTT curves transfers to CGM measurements obtained during an at-home glucose challenge. This setting reflects a plausible path toward scalable metabolic screening, where models can leverage high-quality clinical measurements during development while operating on substantially more accessible wearable measurements at deployment \citep{metwally2025prediction}.

\begin{table}[t]
\setlength{\tabcolsep}{3pt}
\renewcommand{\arraystretch}{0.85}
\caption{Linear probe evaluation for CGM metabolic subphenotyping across three regimes and two endpoints.
All results are mean over 20 seeds.}
\label{tab:cgm}
\centering
\resizebox{0.92\linewidth}{!}{%
\begin{tabular}{clcccccccccccc}
\toprule
& & \multicolumn{6}{c}{AUROC $\uparrow$} & \multicolumn{6}{c}{PRAUC $\uparrow$} \\
\cmidrule(lr){3-8}\cmidrule(lr){9-14}
& & \multicolumn{2}{c}{Home} & \multicolumn{2}{c}{Venous} & \multicolumn{2}{c}{Ven.$\rightarrow$Home} & \multicolumn{2}{c}{Home} & \multicolumn{2}{c}{Venous} & \multicolumn{2}{c}{Ven.$\rightarrow$Home} \\
\cmidrule(lr){3-4}\cmidrule(lr){5-6}\cmidrule(lr){7-8}\cmidrule(lr){9-10}\cmidrule(lr){11-12}\cmidrule(lr){13-14}
& Method & $\beta$-cell & IR & $\beta$-cell & IR & $\beta$-cell & IR & $\beta$-cell & IR & $\beta$-cell & IR & $\beta$-cell & IR \\
\midrule
\sidecite{8}{\citet{muhammad2026cgm}}
  & PCA \citep{shlens2014tutorial}                   & .925 & .821 & .790 & .744 & .927 & .827 & .909 & .842 & .784 & .770 & .911 & .847 \\
  & GluFormer \citep{lutsker2025gluformer}             & .857 & \underline{.889} & .742 & .709 & .801 & .530 & .809 & .880 & .703 & .758 & .756 & .555 \\
  & TS2Vec \citep{yue2022ts2vec}                & .854 & .848 & .735 & .733 & .805 & .713 & .880 & .868 & .716 & .741 & .754 & .720 \\
  & MOMENT$_{\mathrm{S}}$ \citep{goswami2024moment} & .849 & .819 & .712 & .674 & .736 & .830 & .834 & .829 & .668 & .671 & .751 & .852 \\
  & MOMENT$_{\mathrm{L}}$ \citep{goswami2024moment} & .652 & .605 & .683 & .630 & .606 & .713 & .649 & .651 & .604 & .579 & .615 & .750 \\
  & Mantis \citep{feofanov2025mantis}                & .758 & .690 & .674 & .709 & .821 & .766 & .777 & .759 & .634 & .687 & .826 & .809 \\
  & CGM-JEPA \citep{muhammad2026cgm}              & .932 & .842 & \underline{.845} & \underline{.801} & .946 & \underline{.866} & .931 & .872 & .821 & .824 & .941 & \underline{.892} \\
  & X-CGM-JEPA \citep{muhammad2026cgm}            & \underline{.946} & .857 & \textbf{.855} & .800 & \underline{.949} & .865 & \underline{.941} & \underline{.883} & .815 & .822 & \underline{.943} & \underline{.892} \\
\midrule
& TNC \citep{tonekaboni2021unsupervised}                     & .914 & .832 & .831 & \textbf{.805} & .912 & .845 & .909 & .850 & \underline{.831} & \underline{.841} & .882 & .853 \\
& SoftCLT \citep{lee2024soft}                 & .902 & .811 & .735 & .741 & .911 & .831 & .916 & .857 & .717 & .774 & .921 & .872 \\
\midrule
& \method{}                & \textbf{.980} & \textbf{.952} & \textbf{.855} & .796 & \textbf{.978} & \textbf{.952} & \textbf{.972} & \textbf{.946} & \textbf{.882} & \textbf{.848} & \textbf{.967} & \textbf{.943} \\
\bottomrule
\end{tabular}}
\end{table}

\paragraph{Ablations of \method{}} We perform five ablations of \method{} 
(Table~\ref{tab:cgm_ablation}). The most direct test of its core idea 
removes graded weighting: the anchor's own shift becomes the sole positive, 
while all other shifted and unshifted windows serve as uniformly weighted 
negatives. This effectively reduces \method{} to more standard hard contrastive learning with temporal shifts as 
positives. Performance drops 
sharply, most severely under venous$\rightarrow$home transfer ($-17.1$ and 
$-14.2$ AUROC on $\beta$-cell and IR). Temporal shifting alone therefore does not account for 
\method{}'s performance; the gains come from contrasting the shifts of 
similar windows against those of dissimilar ones. Only fixing $\delta=0$, 
which removes temporal shifting entirely, is more damaging, reducing venous 
performance to chance level. The remaining ablations have effects 
concentrated on home and transfer evaluation and particularly on insulin 
resistance. Removing the frame-separation term $H$ or the stop-gradient 
costs up to 6.6 and 7.9 AUROC points, respectively. Computing $c_{ik}$ from
an EMA copy of the encoder trails the 
live encoder by up to 9.0 points. Since the encoder's similarity estimates 
improve throughout training (Figure~\ref{fig:mutualmem}), this suggests 
that the EMA's lag degrades the weights by tying them to an earlier, 
weaker encoder. On venous evaluation, these three variants perform 
comparably to the full model.

\begin{table}[t]
\centering
\setlength{\tabcolsep}{3pt}
\renewcommand{\arraystretch}{0.9}
\caption{Ablation of \method{} components on CGM metabolic subphenotyping. 
The top row reports absolute performance; all other rows report the 
change relative to it. Values are means over 20 seeds.}
\label{tab:cgm_ablation}
\scriptsize
\begin{tabular}{lcccccccccccc}
\toprule
& \multicolumn{6}{c}{AUROC $\uparrow$} & \multicolumn{6}{c}{PRAUC $\uparrow$} \\
\cmidrule(lr){2-7}\cmidrule(lr){8-13}
& \multicolumn{2}{c}{Home} & \multicolumn{2}{c}{Venous} & \multicolumn{2}{c}{Ven.$\rightarrow$Home} & \multicolumn{2}{c}{Home} & \multicolumn{2}{c}{Venous} & \multicolumn{2}{c}{Ven.$\rightarrow$Home} \\
\cmidrule(lr){2-3}\cmidrule(lr){4-5}\cmidrule(lr){6-7}\cmidrule(lr){8-9}\cmidrule(lr){10-11}\cmidrule(lr){12-13}
Variant & $\beta$-cell & IR & $\beta$-cell & IR & $\beta$-cell & IR & $\beta$-cell & IR & $\beta$-cell & IR & $\beta$-cell & IR \\
\midrule
\method{}
  & .980 & .952 & .855 & .796 & .978 & .952
  & .972 & .946 & .882 & .848 & .967 & .943 \\
\midrule
w/o graded weighting
  & \dec{.042} & \dec{.097} & \dec{.049} & \inc{.001} & \dec{.171} & \dec{.142}
  & \dec{.046} & \dec{.090} & \dec{.118} & \dec{.008} & \dec{.159} & \dec{.125} \\
w/o frame separation $H$
  & \dec{.018} & \dec{.066} & \inc{.018} & \inc{.004} & \dec{.012} & \dec{.057}
  & \dec{.020} & \dec{.044} & \dec{.010} & \dec{.015} & \dec{.010} & \dec{.033} \\
w/o stop-gradient
  & \dec{.037} & \dec{.079} & \inc{.019} & \dec{.010} & \dec{.025} & \dec{.069}
  & \dec{.034} & \dec{.052} & \dec{.009} & \dec{.019} & \dec{.025} & \dec{.047} \\
EMA grader ($m{=}0.999$)
  & \dec{.034} & \dec{.090} & \inc{.009} & \inc{.016} & \dec{.032} & \dec{.090}
  & \dec{.031} & \dec{.057} & \dec{.020} & \dec{.009} & \dec{.026} & \dec{.054} \\
$\delta = 0$
  & \dec{.227} & \dec{.198} & \dec{.398} & \dec{.326} & \dec{.220} & \dec{.146}
  & \dec{.265} & \dec{.203} & \dec{.492} & \dec{.404} & \dec{.232} & \dec{.138} \\
\bottomrule
\end{tabular}
\end{table}

\section{Scaling \method{}: A Bite-Sized, SOTA PPG Foundation Model}
\label{sec:fm}

\paragraph{Setup}

We broadly derive our setup from Pulse-PPG \citep{saha2025pulse}, a prior state-of-the-art open PPG foundation model pretrained on field-collected wearable data. We inherit its Mobile Open Observation of Daily Stressors (MOODS) pretraining dataset~\citep{neupane2024momentary}, a large-scale collection of raw smartwatch PPG recorded during participants' everyday lives, as well as its diverse downstream evaluation: PPG-BP~\citep{liang2018new}, WESAD~\citep{schmidt2018introducing}, PPG-DaLiA~\citep{reiss2019deep}, MOODS~\citep{neupane2024momentary}, and SDB~\citep{garde2014development}. The downstream tasks span wearable field, wearable lab, and clinical settings, cover both classification and regression problems such as stress, activity, heart rate, sleep disturbance, blood pressure, and hypertension, and use input windows at a variety of temporal scales.
We strengthen the downstream evaluation protocol for all models. Rather than evaluating on a single subject split with one random seed, we use subject-level five-fold cross-validation repeated across 15 random seeds for every downstream task. We also expand the set of foundation-model baselines to include more recent general time-series and PPG-specific models. These include MOMENT~\citep{goswami2024moment}, Chronos 1 \& 2~\citep{ansari2024chronos, ansari2025chronos}, PaPaGei (both S and P variants)~\citep{pillai2025papagei}, AnyPPG~\citep{nie2026anyppg}, and Pulse-PPG~\citep{saha2025pulse}.

For the \method{} PPG foundation model, we apply a pretraining window length of 1 minute. This contrasts with \cite{saha2025pulse}, where longer, four-minute windows were specifically chosen to capture the multi-minute physiological dynamics associated with stress. \method{}, however, explicitly learns temporal relationships across shifted signal windows during pretraining. We therefore use a shorter window intended to provide a more general representation across tasks operating at different temporal scales. Similar to \cite{saha2025pulse} we use a ResNet backbone, however, we substantially reduce the size of the encoder to 560K parameters, yielding a 96-dim representation. We manually explored a range of architectural and training hyperparameters for the \method{} foundation model, including model depth and width, learning rate, and \method{}-specific parameters. Full details of the \method{} ResNet architecture and pretraining hyperparameters are provided in Appendix~\ref{app:ppgfm}.

\paragraph{Results}
\label{sec:fm-results}
Table~\ref{tab:ppg-fm} summarizes linear-probe performance across 11 downstream classification and regression tasks, spanning five datasets and three settings: wearable field, wearable lab, and clinical. Overall, \method{} is the strongest-performing approach, achieving the best performance on 7 of 11 tasks and outperforming all unimodally pretrained PPG foundation models on 10 of 11. The gains are particularly pronounced for hypertension prediction, where \method{} exceeds the next-best models by roughly 6--7 percentage points in both AUROC and PRAUC.

While \method{} is the best-performing model overall, it is also by far the smallest in our evaluation: the next-smallest, AnyPPG, is over 7$\times$ larger and additionally requires paired PPG and ECG recordings during pretraining. The comparison with Pulse-PPG, the strongest unimodally pretrained baseline (outperforming both PaPaGei variants on 6 of 11 tasks), is especially stark. It is roughly 50$\times$ larger and uses the same pretraining dataset and evaluation splits, yet \method{} outperforms it on 10 of 11 tasks; on the sole exception, sleep, the gap is under one percentage point on both AUROC and PRAUC. Further, attempts at downscaling by \cite{saha2025pulse} show that reducing Pulse-PPG even to $\sim$4.73M parameters causes AUROC losses up to 0.139 in downstream tasks.

On the four tasks where \method{} is not the top performer, two involve marginal differences from the best result in the table, with gaps to the best competing model averaging 1.5 percentage points. The other two cases (DaLiA heart-rate estimation and sleep classification on SDB) are more substantial. For the former, \method{} nevertheless remains the best-performing unimodally pretrained PPG model, with only general time-series foundation models and multimodally pretrained AnyPPG performing better. On sleep, however, \method{} underperforms by roughly six percentage points relative to the best performer, PaPaGei-P. We suspect this reflects a limitation of the MOODS pretraining dataset, which contains no sleep recordings. Consistent with this explanation, Pulse-PPG, a generally strong model trained on the same corpus, also performs poorly on SDB. PaPaGei, by contrast, is pretrained with sleep data and performs best in this setting though even it is hardly better than random. This limitation could be addressed in future work by expanding MOODS to include sleep recordings.

\begin{table}[t]
\setlength{\tabcolsep}{3pt}
\renewcommand{\arraystretch}{0.85}
\caption{Downstream performance of the \method{} PPG foundation model against
general time-series and PPG-specific foundation models. All results are mean over 15 seeds of 5-fold cross-validation. \textbf{Bold}
is best, \underline{underline} second best.}
\label{tab:ppg-fm}
\centering
\resizebox{0.92\linewidth}{!}{%
\begin{tabular}{cllcccccccc}
\toprule
& & & \multicolumn{3}{c}{\textit{General TSFMs}} & \multicolumn{5}{c}{\textit{PPG FMs}} \\
\cmidrule(lr){4-6}\cmidrule(lr){7-11}
& & & MOMENT & Chronos & Chronos-2 & Pulse-PPG & PaPaGei-S & PaPaGei-P & AnyPPG & \method{} \\
\cmidrule(lr){4-11}
& \multicolumn{2}{l}{Params (M)}  & 341.25 & 201.37 & 119.48 & 28.50 & 5.79 & 4.99 & \underline{4.04} & \textbf{0.56} \\
& \multicolumn{2}{l}{Embedding dim} & 1024 & 768 & 768 & 512 & 512 & 512 & 512 & 96 \\
& \multicolumn{2}{l}{Pretraining data} & \textit{Gen.\ TS} & \textit{Gen.\ TS} & \textit{Gen.\ TS} & \textit{PPG} & \textit{PPG} & \textit{PPG} & \textit{PPG+ECG} & \textit{PPG} \\
\midrule
\side{5}{PPG-BP}
  & \multirow{2}{*}{Hypertension} & AUROC $\uparrow$ & \underline{.6749} & .6353 & .6659 & .6487 & .6373 & .6357 & .6739 & \textbf{.7347} \\
  &                               & PRAUC $\uparrow$ & .3951& .3642 & .4103 & .3848 & .3573 & .3567 & \underline{.4250} & \textbf{.4960} \\
  & Systolic BP  & MAE $\downarrow$ & 17.487 & 16.979 & 16.195 & 16.622 & 15.960& \underline{15.470} & 18.429 & \textbf{14.007} \\
  & Diastolic BP & MAE $\downarrow$ & 10.610 & 10.203 & 10.103 & 9.649 & 10.096 & \underline{9.148} & 11.164& \textbf{8.294}\\
  & HR           & MAE $\downarrow$ & 4.950& 5.525 & 4.815 & 4.815 & 5.030 & \underline{4.765} & 5.747 & \textbf{4.433}\\
\midrule
\side{4}{WESAD}
  & \multirow{2}{*}{Stress (2)} & AUROC $\uparrow$ & .8507 & .8642 & .9302 & .9388 & .8383 & .8486 & \underline{.9520} & \textbf{.9630}\\
  &                             & PRAUC $\uparrow$ & .7259 & .7218 & .8675& .8734 & .6537 & .7168 & \underline{.9017} & \textbf{.9387} \\
  & \multirow{2}{*}{Stress (4)} & AUROC $\uparrow$ & .6969& .7394& .7431 & .7197 & .7046 & .6599 & \textbf{.7562} & \underline{.7455} \\
  &                             & PRAUC $\uparrow$ & .4665 & .5174 & \underline{.5425} & .5049 & .4594 & .4299 & \textbf{.5588} & .5404 \\
\midrule
\side{3}{DaLiA}
  & \multirow{2}{*}{Activity} & AUROC $\uparrow$ & .7798 & \textbf{.8069} & .7937 & .7744 & .7164 & .7399 & .7988 & \underline{.7992} \\
  &                           & PRAUC $\uparrow$ & .3250 & \textbf{.3773} & .3461 & .3126 & .2478 & .2821 & \underline{.3641} & .3543\\
  & HR & MAE $\downarrow$ & \underline{8.460} & 8.830 & 9.628& 10.105 & 14.099 & 12.232  & \textbf{6.643}  & 9.790  \\
\midrule
\side{4}{MOODS}
  & \multirow{2}{*}{Activity} & AUROC $\uparrow$ & .7387  & .8325  & .8574  & .8635  & .6917  & .7140  & \underline{.8657}  & \textbf{.8835}  \\
  &                           & PRAUC $\uparrow$ & .3645  & .5001  & .5415  & .5465  & .2982  & .3168  & \underline{.5660}  & \textbf{.6058}  \\
  & \multirow{2}{*}{Stress}   & AUROC $\uparrow$ & .5163  & .5384  & .5360  & .5438  & .5316  & .5526  & \underline{.5556}  & \textbf{.5591}  \\
  &                           & PRAUC $\uparrow$ & .4193  & .4378  & .4292  & .4363  & .4282  & .4391  & \underline{.4464}  & \textbf{.4500}  \\
\midrule
\side{2}{SDB}
  & \multirow{2}{*}{Sleep} & AUROC $\uparrow$ & .5039  & .5364  & \underline{.5425}  & .4967  & .5257  & \textbf{.5507}  & .4929  & .4928  \\
  &                        & PRAUC $\uparrow$ & .3823  & .3931  & \underline{.4017}  & .3632  & .3892  & \textbf{.4118}  & .3727  & .3544  \\
\bottomrule
\end{tabular}}
\end{table}

\section{Conclusion}
We introduce \method{}, an efficient, scalable self-supervised objective for time series that unifies structural and temporal similarity by contrasting the temporal shifts of windows the encoder deems similar against those it deems dissimilar. Across PPG, ECG, EEG, accelerometry, and CGM, \method{} broadly achieves state-of-the-art performance. We further demonstrate it at scale with a 560K-parameter PPG foundation model that achieves state-of-the-art downstream performance while having significantly fewer parameters than the PPG foundation models in our evaluation.

\bibliography{iclr2027_conference}
\bibliographystyle{iclr2027_conference}

\appendix

\section{Multi-Domain Benchmark Details}
\label{app:pulse}
\label{app:pulse_hparams}

All methods share the training-hyperparameter search described in
Section~\ref{sec:pulse} (30 random-search trials), with method-specific
hyperparameters varying independently within the same trials. The best
configuration is selected by validation performance.

\paragraph{\method{}.}
We sweep three method-specific hyperparameters: the offset scale
$\sigma \in \{0.5,\allowbreak 1,\allowbreak 2,\allowbreak 4,\allowbreak 8,\allowbreak 16\}$ (in window lengths), the contrastive
temperature $\tau \in \{0.03,\allowbreak 0.04,\allowbreak 0.05,\allowbreak 0.07,\allowbreak 0.10,\allowbreak 0.15\}$, and the weighting
temperature $\tau_w \in \{0.04,\allowbreak 0.06,\allowbreak 0.10,\allowbreak 0.14,\allowbreak 0.20\}$. The offset is drawn as
$\delta \sim \mathcal{N}(0, \sigma^2)$ and resampled until
$\delta_{\min} \le |\delta| \le \delta_{\max}$, with $\delta_{\min} = 1$ and
$\delta_{\max} = 2.5\sigma$.

\paragraph{SoftCLT.}
Following the classification search space of \citet{lee2024soft}, we sweep the
instance-wise sharpness $\tau_I \in \{1, 2, 3, 4, 5, 10, 20\}$ and the temporal
sharpness $\tau_T \in \{0.5, 1.0, 1.5, 2.0, 2.5\}$, keeping the loss weight fixed
at $\lambda = 0.5$ as in the original work.

\paragraph{TNC.}
We sweep the positive-unlabeled weight $w \in \{0, 0.05, 0.1\}$, where $w = 0$
disables the PU correction, the Monte Carlo sample size $\in \{10, 20, 40\}$, and
the augmentation factor $\in \{5, 7\}$. These ranges span the per-dataset
settings used in the official implementation of
\citet{tonekaboni2021unsupervised}.

\begin{table}[!ht]
\centering
\caption{Final method-specific hyperparameters selected on the validation set for
each domain.}
\label{tab:pulse_final_hparams}
\begin{tabular}{llcccc}
\toprule
Method & Hyperparameter & PPG & ECG & HAR & EEG \\
\midrule
\multirow{3}{*}{\method{}} & $\sigma$            & 1    & 8    & 8    & 2     \\
                           & $\tau$              & 0.03 & 0.04 & 0.03 & 0.04  \\
                           & $\tau_w$            & 0.06 & 0.14 & 0.10 & 0.10  \\
\midrule
\multirow{2}{*}{SoftCLT}   & $\tau_I$            & 10   & 4    & 20   & 5     \\
                           & $\tau_T$            & 0.5  & 1.0  & 2.0  & 2.5   \\
\midrule
\multirow{3}{*}{TNC}       & $w$                 & 0    & 0    & 0.1  & 0.05  \\
                           & MC sample size      & 20   & 10   & 10   & 20    \\
                           & Augmentation factor & 5    & 5    & 5    & 7     \\
\bottomrule
\end{tabular}
\end{table}

\begin{table}[!ht]
\centering
\caption{Selected shared training hyperparameters. Following \citet{chen2025self},
all three methods draw from the same grid (epochs $\in \{50, 100, 200\}$,
learning rate $\in \{10^{-3}, 5{\times}10^{-4}, 10^{-4}\}$, weight decay
$\in \{10^{-3}, 10^{-4}, 10^{-5}\}$, batch size $\in \{64, 128\}$). A batch size of
64 was selected for all methods and domains, except \method{} on HAR, which
selected 128.}
\label{tab:pulse_shared_hparams}
\begin{tabular}{llcccc}
\toprule
Method & Hyperparameter & PPG & ECG & HAR & EEG \\
\midrule
\multirow{3}{*}{\method{}} & Epochs        & 100       & 200       & 50        & 100       \\
                           & Learning rate & $10^{-4}$ & $10^{-4}$ & $10^{-4}$ & $10^{-4}$ \\
                           & Weight decay  & $10^{-5}$ & $10^{-5}$ & $10^{-5}$ & $10^{-4}$ \\
\midrule
\multirow{3}{*}{SoftCLT}   & Epochs        & 200       & 100       & 200       & 200       \\
                           & Learning rate & $5{\times}10^{-4}$ & $10^{-4}$ & $5{\times}10^{-4}$ & $10^{-3}$ \\
                           & Weight decay  & $10^{-4}$ & $10^{-4}$ & $10^{-4}$ & $10^{-4}$ \\
\midrule
\multirow{3}{*}{TNC}       & Epochs        & 100       & 50        & 200       & 100       \\
                           & Learning rate & $10^{-3}$ & $10^{-3}$ & $10^{-3}$ & $5{\times}10^{-4}$ \\
                           & Weight decay  & $10^{-5}$ & $10^{-3}$ & $10^{-3}$ & $10^{-5}$ \\
\bottomrule
\end{tabular}
\end{table}

\clearpage
\section{PPG Foundation Model Details}
\label{app:ppgfm}

Following \citet{saha2025pulse}, the encoder is a 1D ResNet; we reduce it to
560K parameters. Table~\ref{tab:ppgfm_hparams} lists the final configuration.
 
\begin{table}[!ht]
\centering
\caption{PPG foundation model architecture and pretraining hyperparameters.}
\label{tab:ppgfm_hparams}
\begin{tabular}{ll}
\toprule
\multicolumn{2}{l}{\textit{Encoder}} \\
Architecture        & Stem conv + 6 residual blocks \\
Channels            & $48 \to 96$ \\
Kernel size         & 11 \\
Downsampling        & Stride 2 every second block ($8\times$ total) \\
Dropout             & 0.5 \\
Pooling             & Global average \\
Embedding dimension & 96 \\
Parameters          & 560K \\
\midrule
\multicolumn{2}{l}{\textit{Objective}} \\
Window              & 60\,s (3000 samples at 50\,Hz) \\
$\sigma$            & 4 \\
$[\delta_{\min}, \delta_{\max}]$ & $[1, 10]$ windows \\
$\tau$              & 0.03 \\
$\tau_w$            & 0.10 \\
\midrule
\multicolumn{2}{l}{\textit{Optimization}} \\
Optimizer           & Adam, learning rate $10^{-4}$, no weight decay \\
Batch size          & 32 \\
Training            & 4 epochs $\times$ 40{,}000 steps \\
Checkpoint          & Lowest validation loss (epoch 4) \\
\bottomrule
\end{tabular}
\end{table}

\end{document}